%% file: main.tex
\documentclass[preprint,12pt]{elsarticle}

\usepackage{amssymb}
\usepackage{url}
\usepackage{latexsym}
\usepackage{amsmath}
\usepackage{algorithmic}
\usepackage[ruled, vlined, linesnumbered]{algorithm2e}
\usepackage[utf8]{inputenc}
\usepackage{color}
\usepackage{subfigure}
\usepackage{enumitem}
\usepackage{amssymb}
\usepackage{booktabs}
\usepackage{multirow, multicol}
\usepackage{amsmath}
\usepackage{tabularx}
\usepackage{adjustbox}
\usepackage{pifont}
\newcommand{\eat}[1]{}

\usepackage{lineno}

\def\I{\mathcal{I}}

\def\C{\mathcal{C}}
\def\F{\mathcal{F}}
\def\ours{{PGB\xspace}}

\journal{}

\begin{document}

\begin{frontmatter}


\title{PGB: One-Shot Pruning for BERT via Weight Grouping and Permutation}
\author[1]{Hyemin Lim}
\ead{hyemin8670@naver.com}

\author[1]{Jaeyeon Lee}
\ead{dlwodus159@naver.com}

\author[1]{Dong-Wan Choi\corref{cor1}}
\ead{dchoi@inha.ac.kr}

\cortext[cor1]{Corresponding author}
\affiliation[1]{
            organization={Department of Computer Science and Engineering, Inha University},
            addressline={100 Inharo},
            city={Incheon},
            % postcode={22212},
            country={South Korea}}

% \begin{highlights}
% \item 
% \end{highlights}

\input{abstract}
%% Keywords

\end{frontmatter}

% \linenumbers


\input{section1}
\input{section2}

\input{section3}

\input{section4}

\input{section5}

\input{section6}

%% Refer following link for more details.
%% https://en.wikibooks.org/wiki/LaTeX/Mathematics
%% https://en.wikibooks.org/wiki/LaTeX/Advanced_Mathematics

%% Use a table environment to create tables.
%% Refer following link for more details.
%% https://en.wikibooks.org/wiki/LaTeX/Tables

% \bibliography{reference}

\appendix
\input{appendix}



\end{document}

%% file: abstract.tex
\begin{abstract}
Large pretrained language models such as BERT suffer from slow inference and high memory usage, due to their huge size. Recent approaches to compressing BERT rely on iterative pruning and knowledge distillation, which, however, are often too complicated and computationally intensive. This paper proposes a novel semi-structured one-shot pruning method for BERT, called \textit{Permutation and Grouping for BERT} (PGB), which achieves high compression efficiency and sparsity while preserving accuracy. To this end, PGB identifies important groups of individual weights by permutation and prunes all other weights as a structure in both multi-head attention and feed-forward layers. Furthermore, if no important group is formed in a particular layer, PGB drops the entire layer to produce an even more compact model. Our experimental results on BERT$_{\text{BASE}}$ demonstrate that PGB outperforms the state-of-the-art structured pruning methods in terms of computational cost and accuracy preservation.
\end{abstract}

%% file: section1.tex
\section{Introduction}
% Hook(background)
Transformer-based models \cite{Vaswani2017} including BERT \cite{BERT}, RoBERTa \cite{Roberta} and GPT-3 \cite{GPT} have achieved great performance on natural language processing (NLP) tasks. However, all these models suffer from a large number of parameters, which often limits their applications due to high computational cost and memory usage. To overcome this limitation, extensive research has been conducted to reduce the model size of transformer architectures.

Recent works on compressing BERT adopt two primary strategies, pruning \cite{Han} and knowledge distillation (KD) \cite{KD}. Pruning can further be classified into two categories based on how many times pruning and recovery processes are performed: one-shot pruning \cite{SNIP} and iterative pruning \cite{Thinet,LoB}. Even though one-shot pruning is simple and computationally efficient as it conducts only one pruning phase, it tends to be less effective to maintain high accuracy. Therefore, the dominant approach for BERT is taking iterative steps of pruning and recovery while training with original dataset.

Furthermore, recent pruning methods \cite{DynaBERT,block,Xia} attempt to overcome the low pruning performance with the help of KD, which has been successful in maintaining high performance in BERT \cite{DistilB,TinyBERT}. However, the distillation process can be even more time-consuming than iterative pruning, and it is often too complicated to identify what aspects of the teacher model should be matched to the student model, particularly in the BERT architecture.

%figure 1
\begin{figure}[t!]
    \centering
    \includegraphics[width=0.82 \columnwidth]{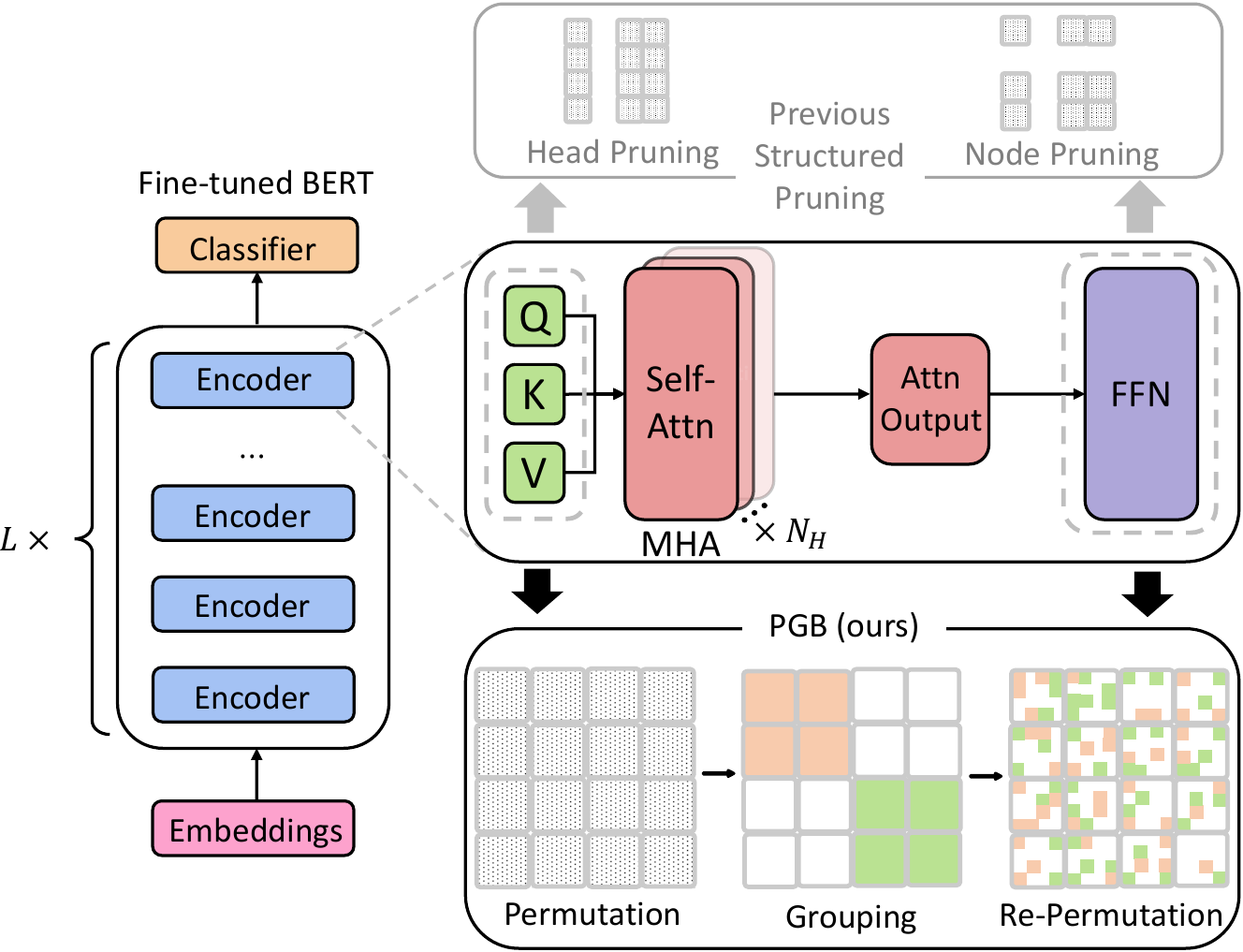}
    \caption{Permutation and Grouping for BERT (PGB), where weight matrices are grouped and all individual weights not belonging to groups are pruned.}
    \label{fig:overview}
\end{figure}

%Motivation
In this paper, we design a novel one-shot pruning method for a task-specific BERT, aiming to achieve both high compression efficiency and high accuracy. Due to the lack of recovery chances, it is quite challenging for one-shot pruning to maintain high performance. Our proposal is to devise a semi-structured pruning method, called \textit{Permutation and Grouping for BERT} (\ours), which combines the benefits of both unstructured and structured pruning. Thus, PGB effectively reduces the model size without sacrificing so much of the accuracy as in unstructured pruning, while ensuring computational efficiency as in structured pruning.

%Methodology Description
Our PGB method is illustrated in Figure \ref{fig:overview}. Basically, we apply a group-based pruning scheme \cite{DGC,Zhao} to each structure of BERT, including multi-head attention (MHA) and feed-forward network (FFN). More specifically, PGB constructs important groups of individual weights for each layer to be preserved and prune all other weights that are not in such a group. Although training with group-based architectures from scratch has been studied with a different purpose \cite{GroupFormer,groupbert}, pruning by weight grouping has never been tackled in the context of transformer-based models. A major challenge is how to preserve the original performance of a given task-specific BERT after pruning unimportant weights. To this end, PGB performs an optimal permutation procedure so that more important weights are clustered as a structure, and adaptively determines the number of important groups for each layer of either MHA or FFN, occasionally dropping the entire layer. Finally, its re-permutation process safely rearranges each weight to its original position. 

%Experiment results
A thorough experimental study is conducted by applying our PGB method to $\text{BERT}_{\text{BASE}}$ \cite{BERT} on the GLUE \cite{GLUE} and SQuAD \cite{SQuAD} benchmarks. Our experimental results show that PGB outperforms the state-of-the-art (SOTA) pruning methods in terms of both efficiency and accuracy.

%% file: section2.tex
\section{Background and Related Works} \label{sec:related}
Pretrained language models \cite{BERT,Roberta,GPT} are mostly based on the transformer architecture \cite{Vaswani2017} due to their effectiveness in various NLP tasks. This section first formally describes the typical transformer architecture and then discusses representative techniques for compressing large language models, namely distillation and pruning.

\subsection{Transformer Architecture}
% MHA
The typical transformer architecture consists of encoder and decoder layers, each of which commonly contains two main components: multi-head attention (MHA) and feed-forward network (FFN). In an MHA layer, there are $N_H$ self-attention heads, and each head $h \in [1, N_H]$ involves the following weight matrices, $W_{h}^{Q}, W_{h}^{K}, W_{h}^{V}, W_{h}^{O} \in \mathbb{R}^{{d} \times\frac{d}{N_H}}$. Then, the final output of the MHA layer is computed as follows:
$$
MHA(X)= \sum_{h = 1}^{N_{H}} Attn_{h}(X),
$$
where $Attn_{h}(X)$ is the output of the standard self-attention unit.

% FFN
The output of MHA is then fed into the corresponding FFN layer, which consists of weight matrices $W^{(1)}\in \mathbb{R}^{d \times d_{ffn}}$, $W^{(2)}\in \mathbb{R}^{d_{ffn} \times d}$, $b^{(1)}\in \mathbb{R}^{d_{ffn}}$ and $b^{(2)}\in \mathbb{R}^{d}$, where $d_{ffn}$ (usually $4\times d$) represents the dimension of the
intermediate hidden features. The output of the FFN layer can be computed as follows:
$$
FFN(A)=\sum_{i=1}^{d_{ffn}}GeLU(AW^{(1)}_{:,i}+b^{(1)})W^{(2)}_{i,:}+b^{(2)},
$$
where $A = MHA(X)$.

In transformer-based models, this same structure is repeatedly defined across multiples layers (e.g., 12 layers in $\text{BERT}_{\text{BASE}}$ \cite{BERT}) and each layer has another multiple heads (e.g., 12 heads in BERT$_{\text{BASE}}$ \cite{BERT}). Consequently, they have numerous trainable parameters, which motivates the NLP community to develop various compression methods for these models.

\subsection{Distillation}
Knowledge distillation (KD) \cite{KD} is a compression technique that trains a lightweight student model to mimic the soft output of a large teacher model, leading to competitive performance on downstream tasks. There are some studies where KD methods have been applied to transformer-based models. For example, DistilBERT \cite{DistilB} transfers knowledge to a student model of a fixed-size through a pre-training phase and an optional fine-tuning process. MiniLM \cite{MiniLM} deeply describes the self-attention information of a task-agnostic teacher model by providing a detailed analysis of the self-attention matrices. Both TinyBERT \cite{TinyBERT} and MobileBERT \cite{MoB} transfer knowledge during pretraining using a layer-by-layer strategy. TinyBERT \cite{TinyBERT} additionally performs distillation during fine-tuning. 

Although KD-based methods are shown to be effective at preserving high accuracy, training a student model can be time-consuming; as reported by CoFi \cite{Xia}, TinyBERT \cite{TinyBERT} requires 350 hours and CoFi \cite{Xia} requires 20 hours for compression. Furthermore, 
it is not trivial to effectively distill the knowledge from a multi-layer teacher model with self-attention information to a student model with fewer layers, which involves the problem of layer selection and different loss functions.

\subsection{Pruning}
Pruning \cite{Han} is another popular compression scheme that removes unimportant weights or structures from the target neural network. Following a massive volume of pruning techniques on convolutional neural networks (CNNs), pruning for transformer family has also been studied, falling into either unstructured or structured pruning.

%Unstructured pruning
\paragraph{Unstructured pruning} 
In unstructured pruning, we remove individual weights that are not important, often aiming to reduce the memory storage itself for the target model while maintaining performance, such as the methods based on \textit{lottery ticket hypothesis} \cite{LoB} and \textit{movement pruning} \cite{Mov}. However, in these unstructured pruning methods, it is difficult and complicated to make satisfactory speedup at inference time, often requiring a special hardware.

%Structured pruning
\paragraph{Structured pruning} 
Structured pruning is a simpler and more efficient way to reduce the model size, where we eliminate some structures of parameters that are considered less significant. In the case of transformer architectures, we can remove redundant attention heads \cite{sixteen,voita,SMP}, entire layers of either MHA or FFN \cite{layer1,layer2}, or neurons along with all their connecting weights of FFNs \cite{earlybert}. However, such a coarse-grained pruning scheme inevitably suffers from severe drop in accuracy. Consequently, recent studies \cite{DynaBERT,block,Xia} have explored the combination of pruning with KD to further enhance the performance of compressed networks. Despite higher performance of these combined approaches, they sacrifice efficiency and simplicity in the compression process itself, as KD typically takes lengthy training times and involves complicated matching procedures.

%Semi-structured pruning
\paragraph{Semi-structured pruning} This paper proposes a semi-structured pruning method for BERT in order to achieve a good balance between efficiency and accuracy. To our best knowledge, \textit{block movement pruning} (BMP) \cite{block} is the only one that can be categorized into semi-structured pruning for transformer family, which proposes a block-based pruning approach for each weight matrix of MHA and FFN layers. 

%Group-based pruning 
\paragraph{Group-based pruning} 
Our method is inspired by a grouping strategy, which has been frequently employed to reduce the computational complexity of CNNs. In this approach, a set of filters or channels are grouped together, and the objective is to minimize connectivity and computation between these groups \cite{DGC,Zhao,deeproot,xie,shufflenet}. These grouped architectures are also incorporated into transformer-based models by a few recent works \cite{GroupFormer,groupbert}. However, they focus on designing more efficient architectures to be trained from scratch, not for compressing a trained task-specific model. This work is the first study on group-based pruning on BERT, aiming to compress the task-specific BERT while maintaining its original accuracy.

%% file: section3.tex
%-------------------------------------------------------------------------------------------
\begin{figure*}[t!]
    \centering
    {\includegraphics[width=\textwidth]{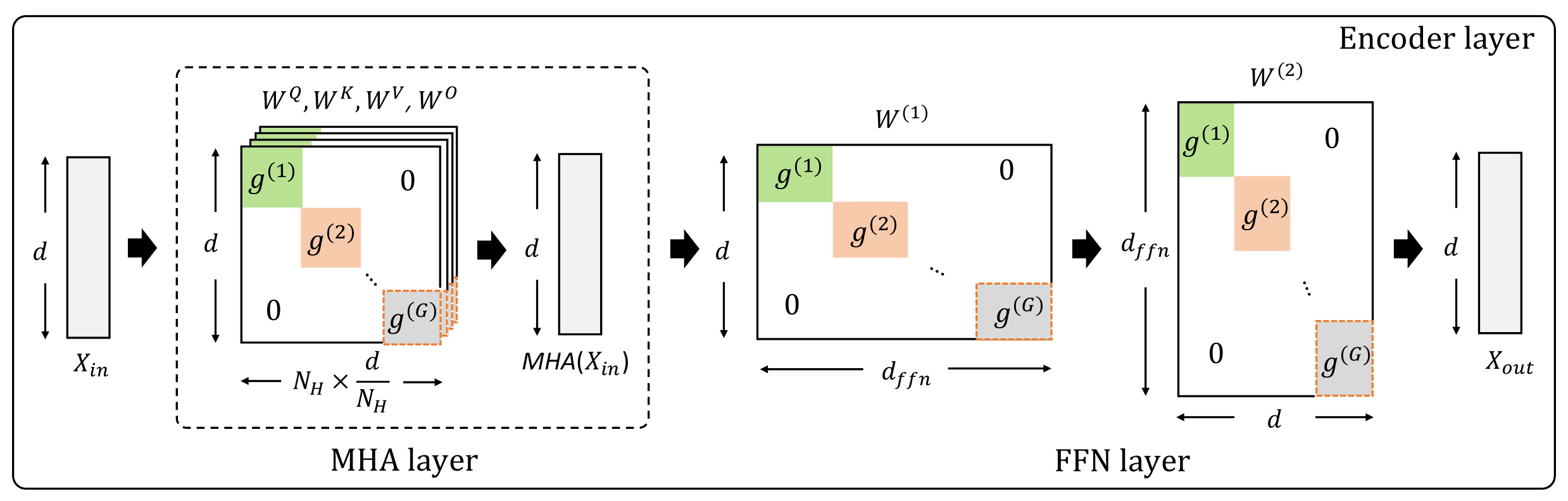}}
    \caption{The grouping procedure of PGB for weight matrices in BERT, which applies to both MHA and FFN layers}
    \label{fig:group}
\end{figure*}

%----------------------------------------------------------------------------------------------

\section{Method} \label{sec:method}
This section presents our one-shot semi-structured pruning method, called \textit{Permutation and Grouping for BERT} (PGB), which applies a grouping procedure to weight matrices of each structure of a given task-specific BERT, as illustrated in Figure \ref{fig:group}.

\subsection{Overall Process} \label{sec:overall}
Unlike existing grouped transformer architectures \cite{GroupFormer,groupbert} that are designed to be trained from scratch, our goal is to find the adaptive grouping for minimizing information loss in the original task-specific BERT. Therefore, as shown in Figure \ref{fig:group}, our PGB method performs the grouped pruning process for each individual weight matrix, rather than partitioning every part of the model into the fixed number of groups as in grouped transformers. In our pruned BERT architectures, the resulting number of groups for each layer can be different, and even a particular layer can be dropped as a whole when no important group is formed in the layer. After the pruning process, we apply the re-permutation step to every pruned weight matrix $\hat{W}$ to restore the original positions of the weights.

\paragraph{Problem formulation}
We first formulate our problem of grouped pruning on BERT. Consider a task-specific BERT $[\Theta_1, ..., \Theta_L]$ with $L$ layers, where $\Theta_i$ consists of weight matrices $W^{Q}, W^{K}, W^{V}, W^{O}$ in its MHA sub-layers, and $W^{(1)}, W^{(2)}$ in its FFN sub-layers. Given a target compression rate $\gamma$ and the task-specific dataset $\mathcal{D}$, our goal is to find the pruned architecture $[\widehat{\Theta}_{1}, ..., \widehat{\Theta}_{L}]$ such that:
\begin{equation}\label{eq:optim}
\begin{matrix}
\min & \sum_{i=1}^{L}  ||\F(\mathcal{D};\Theta_i) - \F(\mathcal{D};\widehat{\Theta}_{i})|| \\
\text{s.t.}& \sum_{i=1}^{L}\C(\widehat{\Theta}_{i}) \approx \gamma \cdot \sum_{i=1}^{L}\C({\Theta_i}),
\end{matrix}
\end{equation}
where $\F(\mathcal{D};\Theta)$ denotes the output of the model parameterized by $\Theta$ for the input data $\mathcal{D}$ and $\C(\cdot)$ is the number of parameters (or FLOPs). As mentioned above, each $\widehat{\Theta}_{i}$ can have different number of groups unlike the equally grouped transformer architectures \cite{GroupFormer,groupbert}.

\paragraph{PGB outline}
Algorithm \ref{alg:pgb} presents how our PGB method finds the group-based pruned architecture for a given task-specific BERT. The algorithm consists of two main phases, namely MHA pruning and FFN pruning. In accordance with previous studies \cite{block,Xia}, our approach focuses on pruning less weights in the MHA sub-layers, compared to those of the FFN sub-layers. Furthermore, since it is crucial for one-shot pruning to deal with a more challenging scenario of pruning a larger number of weights in either MHA or FFN sub-layers, we attempt to minimize the information loss in MHA sub-layers than in FFN sub-layers. To this end, we first take a more conservative approach for pruning weights in MHA sub-layers (Lines 2--5) by trying to find an optimal grouped architecture for each weight matrix $W$ based on its importance, which is performed by the \textsc{Group-Weight-Pruning} subroutine (refer to Algorithm \ref{alg:prune}). Then, we proceed a similar yet more aggressive pruning procedure for the FFN sub-layers (Lines 8--12), where multiple FFN layers that are least important can entirely be dropped in order to meet the remaining budget $C$. Therefore, instead of taking a sequential process across layers, we prioritize the more important layers over the less important ones within the remaining budget $C$.

\begin{algorithm}[t!]
\SetNoFillComment
\DontPrintSemicolon
	{
        \small
		\caption{\small \textsc{PGB-Compression}}\label{alg:pgb}

		\KwIn {$[\Theta_1, ..., \Theta_L]\triangleq$ a given model with $L$ layers, \\ 
                $\gamma\triangleq$ the target compression rate}
		\KwOut {$[\widehat{\Theta}_{1}, ..., \widehat{\Theta}_{L}]~\triangleq~$ the pruned model}

            $C \leftarrow \gamma \cdot \sum_{i=1}^{L}\C({\Theta_i})$ \;
            \tcc{MHA Pruning}
            \For {each layer $i \in [1, L]$}
    		{
                \For {each weight matrix $W$ in MHA of $\Theta_i$}
                {
                    $\widehat{W} \leftarrow$ \textsc{Group-Weight-Pruning}($W$) \;
                    $\widehat{\Theta}_{i}$.\textsc{Append}($\widehat{W}$) \;
                }
    		}
            $C \leftarrow C - \sum_{i=1}^{L}\C({\widehat{\Theta}_{i}})$ \;
            \tcc{FFN Pruning}
            \While{$C > 0$}
            {
                $j \leftarrow$ pick the unused layer with the largest importance score for FFN \;
                \For {each weight matrix $W$ in FFN of $\Theta_j$}
                {
                    $\widehat{W} \leftarrow$ \textsc{Group-Weight-Pruning}($W$) \;
                    $\widehat{\Theta}_{j}$.\textsc{Append}($\widehat{W}$) \;
                    $C \leftarrow C - \C({\widehat{W}})$\;
                }
    
        }
        \Return{$[\widehat{\Theta}_{1}, ..., \widehat{\Theta}_{L}]$}

 }
\end{algorithm}

\subsection{Grouped Weight Pruning}

In order to find the optimal grouping for each weight matrix of BERT, we adapt the technique of \textit{group convolution} pruning \cite{DGC,Zhao}, which is originally intended to prune filters in CNNs, not individual weights as in our problem setting.

The process of our grouped weight pruning is presented in Algorithm \ref{alg:prune}. For each weight matrix $W\in \mathbb{R}^{M\times N}$ in BERT, our method first adaptively determines the number $G$ of groups by measuring the degree of importance of $W$, and drop the entire matrix if the corresponding importance measure is not sufficiently high (Lines 1--3). Then, we permute the matrix to form groups of important weights into block diagonal matrices $g^{(i)}$'s (Line 6). Once the permuted matrix $\widetilde{W}$ is obtained, we extract only the top-left corner block of $\widetilde{W}$ (i.e., $\widetilde{W}[1:\frac{M}{G}, 1:\frac{N}{G}]$) to form a group matrix $g^{(i)}$ (Line 7), and discard all the weights in the region $\widetilde{W}[1:\frac{M}{G},~]$ and $\widetilde{W}[~, 1:\frac{N}{G}]$ from $\widetilde{W}$ (Line 8). The final $\widehat{W}$ would be represented as follows:

    \begin{equation}\label{eq:grouped}
    \begin{aligned}
        \widehat{W} &=\begin{vmatrix}
        g^{(1)}& 0 & \cdots & 0 \\
        0 & g^{(2)} & \cdots  & 0 \\
        \vdots & \vdots  & \ddots  & \vdots  \\
        0 & 0 & \cdots  & g^{(G)} \\
        \end{vmatrix}.
    \end{aligned}
    \end{equation}

\begin{algorithm}[t!]
\DontPrintSemicolon
	{
 
        \small
		\caption{\small \textsc{Grouped-Weight-Pruning}}\label{alg:prune}
     
		\KwIn {$W\triangleq$ a weight matrix of $M \times N$}
		\KwOut {$\widehat{W}~\triangleq~$ the pruned weight matrix}
        $G \leftarrow$ \textsc{Determine-Group-Numbers}($W$)\;
        \If{$G = 0$}{\Return{null}\;}
  
		$\widetilde{W}\leftarrow W$; $~~\widehat{W}\leftarrow$ \textit{null}\;
        \For{each group $i \in [1, G]$}
		{

                $\widetilde{W}\leftarrow$ \textsc{Permutation}($\widetilde{W}$)\;
			    $g^{(i)} \leftarrow \widetilde{W}[1:\frac{M}{G}~,~~1:\frac{N}{G}]$\;
                $\widetilde{W} \leftarrow \widetilde{W}[\frac{M}{G}:~,~~\frac{N}{G}:~]$\;
                $\widehat{W}$.\textsc{Append-Diagonal-Block}($g^{(i)}$)\;
            
		}
        \Return{$\widehat{W}$}\;
    
 }
\end{algorithm}

\paragraph{Finding the optimal permutation}
The permutation procedure (Line 6 in Algorithm \ref{alg:prune}) returns the optimal arrangement $\widetilde{W}$ of individual weights within the given matrix $W \in \mathbb{R}^{M\times N}$. The objective is to cluster more important weights together, forming a group located at the top-left corner block of $\widetilde{W}$. To this end, we determine the optimal pair of permutation vectors $\pi_r$ and $\pi_c$ for the rows and columns of $W$, respectively, that are used to rearrange the weights of $W$, resulting in the formation of $\widetilde{W}$ as follows:
\begin{equation}\label{eq:wtilde}
\begin{matrix}
\max\limits_{\pi_{r},\pi_{c}}& \I(\widetilde{W}[1:\frac{M}{G}, 1:\frac{N}{G}]) \\
\text{s.t.} & \widetilde{W} = W_{\pi_r, \pi_c},
\end{matrix}
\end{equation}
where $\I(\cdot)$ is the total importance score of a given weight matrix and $W_{\pi_r, \pi_c}$ is the resulting matrix when permuting the rows and columns of $W$ using $\pi_r$ and $\pi_c$, respectively. We calculate the importance scores of weights using the second-order information \cite{Brain,second-order, WoodFisher}, which allows us to quantify relative significance of the weights. The following example shows such optimal permutations for a matrix $W \in \mathbb{R}^{4\times 4}$ and $G=2$, assuming that each number in the matrix indicates the importance score of the corresponding weight:
$$
    \begin{vmatrix}
   1 & 0 & 0 & 2 \\
   0 & 1 & 1 & 0 \\     
    2 & 0 & 0 & 1  \\
    0 & 1 & 1 & 0  \\
    \end{vmatrix}  \xrightarrow[\pi_{c}={[1,4,2,3]}]{\pi_{r}=[1,3,2,4]} 
    \begin{vmatrix}
   1 & 2 & 0 & 0 \\
   2 & 1 & 0 & 0 \\     
    0 & 0 & 1 & 1  \\
    0 & 0 & 1 & 1  \\
    \end{vmatrix}.
$$

\paragraph{Heuristic solution for permutation}
Since weight matrices in BERT are typically high-dimensional, we employ an efficient heuristic algorithm \cite{Zhao} that finds sub-optimal permutation vectors. This algorithm alternatively sorts rows or columns based on the summation of importance scores corresponding to the weights of either row vectors within the region $\widetilde{W}[1:\frac{M}{G},~]$ or column vectors within the region $\widetilde{W}[~, 1:\frac{N}{G}]$. Since each sorting process for the rows or the columns changes the arrangement of the corresponding columns or rows, respectively, we repeat this pairwise sorting process a few times (e.g., 6 times in our experiments using BERT$_{\text{BASE}}$).

\paragraph{Adaptive group numbers}
The key property of our method is the adaptive determination of the number $G$ of groups (Line 1 in Algorithm \ref{alg:prune}), based on the importance of weights within $W \in \mathbb{R}^{M \times N}$. Basically, as the number of important weights in $W$ increases, we decrease $G$ and prune a smaller number of weights, whereas if $W$ contains fewer important weights, we increase $G$ to prune a greater number of weights. To this end, we devise the following strategy to adjust $G$ based on the count of weights whose important scores exceed a specified threshold $\tau$:
\begin{enumerate}
    \item Determine the count $n_{\tau}$ of weights in $W$ with importance scores higher than $\tau$.
    \item If $\frac{M\times N}{n_{\tau}} > G_{max}$, then prune the entire $W$.
    \item Otherwise, set $G$ to a value less than $\frac{M\times N}{n_{\tau}}$.
\end{enumerate}
The term $\frac{M\times N}{n_{\tau}}$ is derived from the fact that the number of parameters of a $G$-grouped matrix $\widehat{W}$ is equal to that of $W$ divided by $G$, i.e., $\frac{M\times N}{G}$. Therefore, to ideally cover all $n_{\tau}$ weights, $\widehat{W}$ should have at most $\frac{M\times N}{n_{\tau}}$ groups. Also, we introduce the hyperparameter $G_{max}$ to prevent the formation of excessive groups for non-critical weight matrices (e.g., $G_{max}=6$ in our experiments).

\subsection{Re-Permutation} \label{sec:repermute}

PGB finally performs the re-permutation procedure on every pruned weight matrix $\widehat{W}$ in all the layers of the model to identify the positions of the weights that correspond to the original model. This yields a re-permuted weight matrix $\widehat{W}^{*}$, wherein each weight is returned to its original positions. In this process,  we utilize the permutation vectors $\pi_{r}$ and $\pi_{c}$ that have been stored, and proceed the following operation: 
\begin{equation*}
\begin{aligned}
\widehat{W} \xrightarrow[argsort(\pi_{c})]{argsort(\pi_{r})} \widehat{W}^{*},
\end{aligned}
\end{equation*}
where $argsort(\pi)$ returns the corresponding re-permutation vectors that rearrange the shuffled weights back to their original positions. Note that the final $\widehat{W}^{*}$ after re-permutation is in the same form resulting from fine-grained unstructured pruning, but actual computation at inference time is efficiently performed only with each $g^{(i)} \subseteq \widehat{W}$ as in grouped transformer architectures \cite{GroupFormer,groupbert}.

\paragraph{Weight compensation}
To further restore the performance of the original task-specific BERT model, we update each unpruned weight in every $\widehat{W}^{*}$ by minimizing the following reconstruction error:
\begin{equation}\label{eq:reconstruction}
     \min \left\|\F(X;\widehat{W}^{*})-\F(X;W)\right\|_{2}^{2},
\end{equation}
where $\F(X; W)$ denotes the outputs for a sample dataset $X$.

%retraining
\paragraph{Re-Finetuning}
After all these steps, we perform re-finetuning in the same way as in the original BERT \cite{BERT} to recover the performance that is lost due to the pruning process.

%Experiment result
\begin{table*}
\adjustbox{width= \textwidth}{
\begin{tabular}{clcccccccccccc}

\noalign{\smallskip}\noalign{\smallskip}\toprule
{Pruning}& \multirow{2}{*}{Method}& \multirow{2}{*}{\# Param}  & QNLI & QQP & SST-2 & CoLA & STS-B & MRPC & RTE & $\text{SQuAD}_{1.1}$ & $\text{SQuAD}_{2.0}$ \\ %\cline{3-11}
Ratio&   &     & Acc. & Acc. & Acc. & Mcc. & Spearman & Acc. & Acc.&  EM/F1 & EM/F1   \\
\hline \hline  
0\% &$\text{BERT}_{\text{BASE}}$ & 85M & 91.4 & 91.5 & 93.2 & 58.9 & 89.2 & 86.3 & 66.8 & 80.8/88.3 & 70.9/74.2 \\
\hline 
% BMP$_{50\%}$  & 42.8M & - & 89.4 & 90.3 & 90.7  & - & - & -  & - & -   \\
% BMP$_{88\%}$ & 10.9M & - & 83.2 & 88.9 & 89.3 & - &  - & - & - & -  \\
\multirow{4}{*} {50\%}&EBERT  & 42M  & 89.9 & 90.6 & 90.8 & N.A. & 87.1  &  72.8 & 52.7 & 76.7/85.2 & 68.6/72.5 \\
                    &DynaBERT  & 42.8M  & 88.3 & 90.7 & 91.6& 51.2 & 86.4 & 77.8  & 63.5 & - &  - \\
                    &CoFi & 42.3M  & 88.8 & 90.6 & 90.1 & 53.6 & 88.0 & 83.5 & 56.7 & - & - \\
                    &\textbf {PGB (Ours)} & 42.5M & \textbf{90.3}& \textbf{91.1} & \textbf{92.3} & \textbf{54.9} & \textbf{88.8} & \textbf{84.3}  & \textbf{64.6}  & \textbf{78.0}/\textbf{86.8} & \textbf{69.6}/\textbf{73.5}\\
\hline
\multirow{4}{*} {88\%} &EBERT & 10.9M  & 81.8 & 88.1 & 87.5  & N.A. &  84.9 & N.A. & 49.5 & N.A. & N.A. \\
                    &DynaBERT & 10.7M  & 83.5 & 86.8 & 88.5 & 18.7 &  82.9 & 72.6 & 53.1 & - & -  \\
                    &CoFi & 10.4M  & 84.7 & 89.8 & 89.0 & 32.1 & 85.1 & 75.3 & 52.4 & - & - \\
                    &\textbf{PGB (Ours)} & 10.2M & \textbf{86.4} & \textbf{90.1} & \textbf{89.6}& \textbf{39.5} & \textbf{85.3} & \textbf{78.2} & \textbf{54.3}  & \textbf{71.5}/\textbf{81.2} & \textbf{65.9}/\textbf{69.7} \\
\bottomrule
\end{tabular}
}
\caption{Performance comparison with structured pruning methods using 50\% and 88\% pruning rates on $\text{BERT}_{\text{BASE}}$, where N.A. indicates that the respective method do not achieve the specified level of sparsity.}
\label{tab:comACC}
\end{table*}

%-------------------------------------------------------------------------------------------------
\begin{figure*}[t!]
    \centering
    {\includegraphics[width= \textwidth]{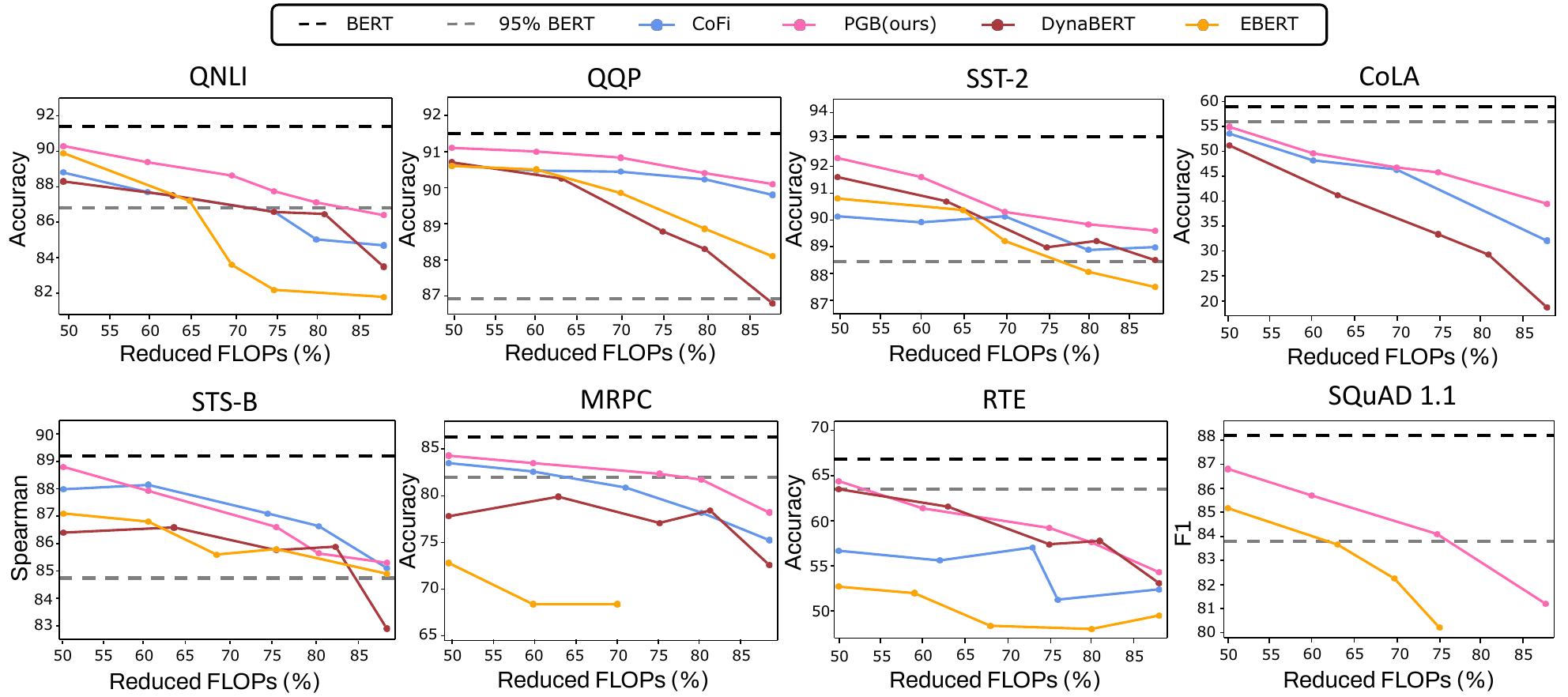}}
    %\subfigure[\label{fig:comacc:b} Accuracy v.s. FLOPs]{\includegraphics[width=1\textwidth,height=4cm]{EMNLP 2022/graph_FLOPs.pdf}}
    \caption{Performance comparison with structured pruning methods varying the reduced FLOPs ratio on $\text{BERT}_{\text{BASE}}$.}
    \label{fig:comacc}
\end{figure*}

%% file: section4.tex
\begin{algorithm}[h]
\SetNoFillComment
\DontPrintSemicolon
	{
		\caption{\small \textsc{PGB-Linear}}\label{alg:infer}

		\KwIn { $X \triangleq$ input X of $S \times M$,\\ 
  $\widehat{W} \triangleq$ a pruned weight matrix with $G$ diagonal groups, where each group is of $\frac{M}{G}\times \frac{N}{G}$,\\
  $\pi_r, \pi_c \triangleq$  permutation vectors for the rows and columns of each weight matrix $W$ }
        \For {each linear operation $X \widehat{W}^*$ s.t. $\widehat{W}^* \xleftarrow[argsort(\pi_{c})]{argsort(\pi_{r})} \widehat{W}$}
		{
            $\widetilde{X} \xleftarrow[\pi_c]{} X$\;
            
            \For {each group $j \in [1,G]$}
            {
                
                $g^{(j)} \leftarrow \widehat{W}[\frac{M}{G}(j-1):\frac{M}{G}j~,\frac{N}{G}(j-1):\frac{N}{G}j]$\;
                $\widetilde{X}^{(j)} \leftarrow \widetilde{X}[1:S~,~~\frac{M}{G}(j-1):\frac{M}{G}j]$\;
                $o^{(j)} \leftarrow \widetilde{X}^{(j)}g^{(j)}$\;
            }

            $O \leftarrow \textsc{Concat}[o^{(1)},o^{(2)},\dots,o^{(G)}]$\;
            $O^{*} \xleftarrow[]{argsort(\pi_r)} O$\;
            \Return ${O^{*}}$\;
		}
        
    }
\end{algorithm}

\section{Cost Analysis for Inference with PGB} \label{sec:cost}
As mentioned above, we make inference using only $G$ groups of each pruned weight matrix $\widehat{W} \in \mathbb{R}^{M\times N}$ for each linear operation $X \widehat{W}^*$, where $X$ is the $S$-length input sequence. By using this inference module, we ensure $G$ times faster efficiency by reducing the previous time cost $S  \cdot  M \cdot N$ to $S \cdot G  \cdot \frac{M}{G} \cdot \frac{N}{G}$. 

\noindent Algorithm \ref{alg:infer} provides a detailed outline of the linear operation process used at inference time with a pruned model resulting from the \textsc{Group-Weight-Pruning} procedure in Algorithm \ref{alg:prune}. It takes the input sequence $X \in \mathbb{R}^{S \times M}$, the grouped weight matrix $\widehat{W}$ obtained after PGB pruning, and the row and column permutation vectors $\pi_r$ and $\pi_c$ that have been determined during the pruning process. At this point, we take only the groups of remaining weights, $g^{(1)},...,g^{(G)}$ where each $g^{(i)} \in \mathbb{R}^{\frac{M}{G}\times \frac{N}{G}}$. Our inference process first starts with permuting the input $X$ using $\pi_c$ at each linear operation (Line 2). Subsequently, for the pruned weight matrix $\widehat{W}$ and the permuted matrix $\widetilde{X}$, we obtain the output $o^{(j)}$ for each group with its corresponding weight matrix $g^{(j)}$ and permuted input matrix $\widetilde{X}^{(j)}$ (Lines 3--6). Once the outputs $o^{(1)},...,o^{(G)}$ for all groups are obtained, we concatenate them to form the output $O$, and permute the output $O$ with respect to $argsort(\pi_r)$ to obtain the final desired output $O^{*} \in \mathbb{R}^{S \times N}$ (Lines 7--8), where $argsort(\pi_r)$ returns the permutation vector of indices that rearrange the output values back to their original positions. As mentioned in Section \ref{sec:cost}, \textsc{PGB-Linear} incurs $1/G$ times the cost of the original linear operation, i.e., $X \widehat{W}^*$.

%% file: section5.tex
\section{Experiments}
\subsection{Environment}
\paragraph{Datasets}

We conduct our experiments using two benchmark datasets, $\text{SQuAD}$ \cite{SQuAD,SQuAD2} and seven tasks (QNLI, QQP, SST-2, CoLA, STS-B, MRPC, and RTE) in GLUE \cite{GLUE} on $\text{BERT}_{\text{BASE}}$ \cite{BERT}.
In our experiments, we use 2K samples from the training data for each benchmark dataset. Also, we perform re-finetuning on the pruned model using the training data of each NLP downstream task and evaluate on the corresponding dev dataset for each task. For detailed information on the benchmarks, please refer to \ref{app:gluedetail}. 

\paragraph{Compared methods}
In order to evaluate the performance of PGB, we conduct comparative experiments with the state-of-the-art structured pruning methods for BERT, including EBERT \cite{ebert}, DynaBERT \cite{DynaBERT}, and CoFi \cite{Xia}. In order to examine their pruning performance, we train each method without using distillation and data augmentation, both of which can commonly be applied to each pruning method.

\paragraph{Implementation details} 
Our PGB method is implemented using Python 3.7.15, PyTorch \cite{Pytorch} and CUDA 11.6, along with the Huggingface library \cite{Wolf}, which incorporates the latest NLP techniques. We set the hyperparmeter $N_{perm}$ that controls the number of sorting operations in the permutation step, and $G_{max}$ that indicates the maximum number of groups, ensuring that the size of the grouped model is within the given budget capacity. All the experiments are conducted on a PC with NVIDIA GeForce RTX A6000. We report that all experimental results are the average of 5 random seeds within a range between 0 and 10.

% Baseline model
As our target model, we use fine-tuned BERT \cite{BERT} for specific tasks. Even though we mainly compare the performance using $\text{BERT}_{\text{BASE}}$, we also conduct experiments using $\text{RoBERTa}_{\text{BASE}}$ and $\text{DistilBERT}_{\text{BASE}}$ \cite{DistilB} (refer to \ref{app:roberta} and \ref{app:distil}). Full experimental details can be found at \ref{app:exdetail}.

\subsection{Main Experimental Results}

\paragraph{Performance comparison}
Table \ref{tab:comACC} and Figure \ref{fig:comacc} show the performance comparison results of PGB with prior structured methods, using the seven tasks of GLUE and SQuAD. To equalize the resulting size of pruned models, we re-implement the released code of the compared methods by ourselves. For CoFi and DynaBERT, since there is no publicly available code for SQuAD, comparative experiments for these two methods were conducted exclusively on the GLUE benchmarks. Table \ref{tab:comACC} demonstrates the corresponding results with pruning rates 50\% and 88\%, where 88\% is the maximum pruning rate that can be achieved by all the compared methods. It is clearly observed that the proposed PGB method outperforms the previous structured pruning methods in all task-specific BERT models, which indicates that PGB is not only the fastest pruning method as observed in Table \ref{time}, but also highly effective to preserve the information of the original model. This empirically implies that we can properly compress transformer architectures with one-shot pruning, without relying on complicated and time-consuming methods.

Figure \ref{fig:comacc} shows how the compression performance changes when varying the target model size in terms of the number of FLOPs. We can observe that PGB generally achieves the best performance among the compared methods. More importantly, PGB tends to work better at higher compression rates, as the performance degradation of PGB gets smaller than those of the other methods as the size of compressed model decreases. These results imply that PGB is capable of maintaining high performance even at extreme compression cases. Considering the compression speed of PGB, we can claim that PGB is a computationally efficient yet accurate compression method for transformer architectures. 

\begin{table}[t!]
\begin{tabular}{clccccc}
\toprule
\multirow{2}{*}{ Pruning ratio} &\multirow{2}{*}{Method}& QNLI & QQP & SST-2 \\
            &       & Acc. & Acc. & Acc.   \\
\hline \hline
0\% & $\text{BERT}_{\text{BASE}}$ & 91.43 & 91.50 & 93.16   \\
\hline
\multirow{4}{*} {50\%} &\multirow{1}{*}{BMP} &  \multirow{1}{*}{89.40} & \multirow{1}{*}{90.30} & \multirow{1}{*}{90.71}    \\
 & \multirow{1}{*}{LayerDrop}   & \multirow{1}{*}{87.60} & \multirow{1}{*}{90.40} & \multirow{1}{*}{90.30}  \\
& \multirow{1}{*}{SNIP}         & \multirow{1}{*}{89.50} & \multirow{1}{*}{88.90} & \multirow{1}{*}{91.80} \\
& \multirow{1}{*}{$\textbf{PGB}$ (Ours)}    & \multirow{1}{*}{\textbf{90.34}} & \multirow{1}{*}{\textbf{91.06}} & \multirow{1}{*}{\textbf{92.30}} \\
\hline
\multirow{2}{*}{80\%}  &\multirow{1}{*}{BMP }   & \multirow{1}{*}{86.40} & \multirow{1}{*}{89.30} &\multirow{1}{*}{89.79}   \\
    & \multirow{1}{*}{\textbf{PGB} (Ours)}  &  \multirow{1}{*}{\textbf{87.12}} & \multirow{1}{*}{\textbf{90.40}} & \multirow{1}{*}{\textbf{89.84}}       \\
\bottomrule
\end{tabular}
\centering
\caption{Performance comparison with other SOTA pruning methods on $\text{BERT}_{\text{BASE}}$.}
\label{tab:more}
\end{table}

%----------------------------------------------------------------------------
\begin{table}
\centering
\begin{tabular}{c|cc|cc}
\toprule
\multirow{2}{*}{Task} & \multicolumn{2}{c|}{40\%}  & \multicolumn{2}{c}{60\%} \\ \cmidrule(r){2-5}
                    & Vanilla  & Re-finetuned & Vanilla & Re-finetuned  \\
\hline

 SST-2 &  92.2  &  92.7  &90.3 & 91.6 \\

QNLI & 90.1  &  91.0   & 88.2&  89.4  \\

 MRPC &  84.8    &  85.1 & 80.9 & 83.5 \\
\bottomrule
\end{tabular}
\centering
\caption{Ablation study in PGB (ours) at 40\% and 60\% of reduced FLOPs before and after re-finetuning using GLUE.}
\label{tab:vanilla}
\end{table}
%-------------------------------------------------------------------------------------------
\paragraph{Comparison with other SOTA methods}
Table \ref{tab:more} presents additional comparisons between PGB and other state-of-the-art (SOTA) pruning methods, using the reported results from their respective experiments. The compared methods include: hybrid-based block movement pruning (BMP) \cite{block}, LayerDrop that drops certain layers of the model \cite{layer2}, and SNIP that prunes redundant mappings in residual modules \cite{TFsnip}. Similar to the findings in Table \ref{tab:comACC}, PGB shows minimum performance degradation. Of particular interest is the comparison with BMP \cite{block}, as it is introduced as another semi-structured pruning method in Section \ref{sec:related}. PGB turns out to outperform BMP even though the reported results of BMP are not from its fully semi-structured pruning scheme, but rather its improved hybrid version yet without distillation.
\begin{figure}[t!]
    \centering
    {\includegraphics[width=0.7\columnwidth]{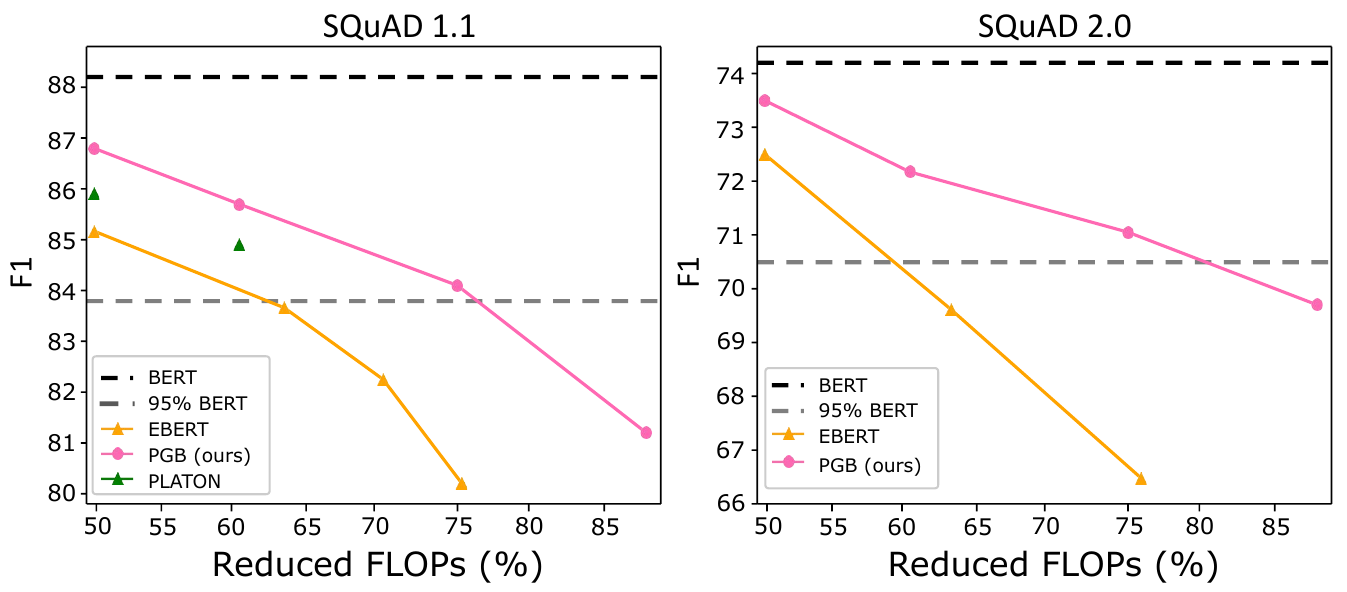}}
    \caption{Performance comparison with structured pruning methods varying the Reduced FLOPs ratio on $\text{SQuAD}$ benchmarks.}
    \label{fig:squadacc}
\end{figure}
\begin{table*}[t!]
\centering
\begin{tabular}{l|c|c|c|c}
%\noalign{\smallskip}\noalign{\smallskip}\hline
\toprule
\multirow{2}{*}{Method} & \multicolumn{2}{c|}{Time for Pruning + Re-finetuning}  & \multicolumn{1}{c|}{FLOPs} & \multicolumn{1}{c}{Accuracy} \\ \cmidrule(r){2-5}
 & (epochs)  &  (hours) & (G) & (\%) \\ 
\hline
EBERT \cite{ebert}      &  3 + 3       & $\leq$ 3.5 + 2        &    2.78    & 88.1 \\
DynaBERT \cite{DynaBERT}&  2 + 3        & $\leq$ 25 + 38        &    \textbf{2.60}   &   86.8  \\
CoFi \cite{Xia}     & 20 + 20           & $\leq$ 26 + 23        &    2.97    & 89.8 \\ \hline
 \textbf{PGB (ours)}& \textbf{0 + 3}  & $\leq$ \textbf{0.1 + 2} &    \textbf{2.60}  & \textbf{90.1} \\
%\hline
\bottomrule
\end{tabular}
\caption{Comparison of the pruning efficiency using QQP with 88\% pruning rate on $\text{BERT}_{\text{BASE}}$.}
\label{time}
\end{table*}
\paragraph{Experiments on SQuAD Benchmarks}\label{app:squad} 

We compare additional pruning methods with PGB for $\text{SQuAD}_{v1.1}$ \cite{SQuAD} and $\text{SQuAD}_{v2.0}$ \cite{SQuAD2}. The compared methods exclude knowledge distillation (KD) and data augmentation. Specifically, our approach is compared with PLATON \cite{platon}, representing the state-of-the-art in unstructured pruning, and EBERT \cite{ebert}. In this context, PLATON utilized the extended $\text{PLATON}_{\text{structure}}$ based on structured pruning. Our PGB method shows minimal performance degradation compared to other methods, as shown in Figure \ref{fig:squadacc}.

%Roberta result
\begin{figure*}[t!]
    \centering
    {\includegraphics[width=  \textwidth]{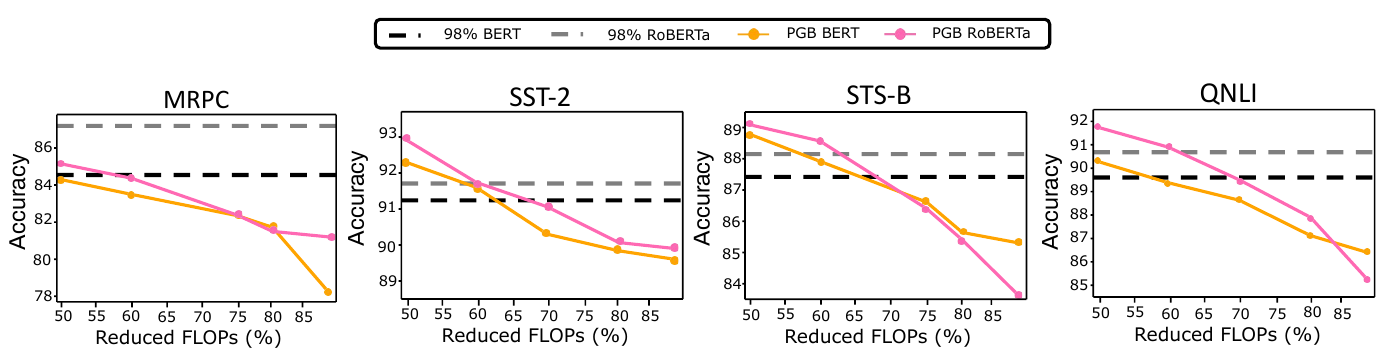}}
    \caption{PGB with $\text{BERT}_{\text{BASE}}$ and $\text{RoBERTa}_{\text{BASE}}$ }
    \label{fig:roberta}
\end{figure*}

\paragraph{Efficiency of pruning procedures}
To evaluate the efficiency of each pruning method, we measure how long each method takes to prune the original model and re-finetune the pruned model, as shown in Table \ref{time}. In particular, we report the case of pruning with 88\% on QQP, where the time cost was maximized. Thanks to its one-shot pruning scheme, our PGB method is clearly the fastest pruning method among all the compared methods, while maintaining the best accuracy after pruning and re-finetuning. PGB takes at most 2.1 hours to obtain the final compressed model, whereas most of the other methods takes more than a day to get the same-sized model. Table \ref{time} also reports the computational cost of each compressed model in terms of the number of FLOPs. Although PGB is a semi-structured pruning method, it manages to achieve a comparable model efficiency, without any specific hardware support, to those of the fully structured pruning methods, EBERT \cite{ebert}, DynaBERT \cite{DynaBERT} and CoFi \cite{Xia}.

\subsection{More Experimental Results}
\paragraph{Pruning on DistilBERT}\label{app:distil}
We conduct additional experiments on PGB with pruning rates of 40\% and 60\% for $\text{DistilBERT}_{\text{BASE}}$, and the results are presented in Table \ref{tab:distil}. While $\text{DistilBERT}_{\text{BASE}}$ is half the size of $\text{BERT}_{\text{BASE}}$, as indicated in Table \ref{tab:BERT}, the model configuration remains the same, excluding the number of layers. Consequently, the values of permutation ($N_{perm}$) and the maximum group number ($G_{max}$) are identical. The experimental results in Table \ref{tab:distil} demonstrate that even for a smaller model, our grouping-based pruning scheme does not result in substantial information degradation.

\begin{table} [t!]
\begin{tabular}{cccccc}
\noalign{\smallskip}\noalign{\smallskip}\hline

Pruning Ratio & Method & QNLI &SST-2 &MRPC & STS-B   \\ 

\hline \hline
{0\%}& $\text{DistilBERT}_{\text{BASE}}$ & \multirow{1}{*}{88.6} & \multirow{1}{*}{91.3} & \multirow{1}{*}{84.8} &\multirow{1}{*}{85.8} \\

{40\%}  &\multirow{2}{*}{PGB (ours)}     &  \multirow{1}{*}{88.1}  & \multirow{1}{*}{91.2} & \multirow{1}{*}{84.6}  & \multirow{1}{*}{85.5}  \\

{60\%}  &         & \multirow{1}{*}{86.5}  & \multirow{1}{*}{89.3} &\multirow{1}{*}{83.8}& \multirow{1}{*}{84.3}  \\

\hline
\end{tabular}
\centering
\caption{Performance of PGB on $\text{DistilBERT}_{\text{BASE}}$ using 40\% and 60\% pruning rates}
\label{tab:distil}
\end{table}

\paragraph{Pruning on RoBERTa}\label{app:roberta}
Figure \ref{fig:roberta} shows the proposed PGB results with $\text{RoBERTa}_{\text{BASE}}$. Similar to BERT, PGB with pruning rates of 60\% on RoBERTa is able to  maintain over 98\% of the performance of the original model. When the Pruning rate surpasses 60\%, RoBERTa demonstrates better performance compared to BERT. However, as the sparsity ratio increases, the performance of BERT surpasses RoBERTa.

\subsection{Ablation Studies}\label{app:ablation}
\paragraph{Performance of PGB without re-finetuning}

Table \ref{tab:vanilla} presents the performance comparison between vanilla PGB, which does not perform re-finetuning, and PGB with re-finetuning. We measure the accuracy of the compressed model with 40\% and 60\%  of reduced FLOPs before and after re-finetuning on task-specific BERT models. We can observe that vanilla PGB itself is already effective to maintain high performance after pruning. For the SST-2 and QNLI tasks, the performance of vanilla PGB is either matches or surpasses that of other pruning methods shown in Figure \ref{fig:comacc}. This demonstrates the capability of our grouped pruning operation to maintain the original performance of large transformer-based models with just a single round of pruning.

\begin{figure}[t!]
    \centering
    {\includegraphics[width=0.7\columnwidth]{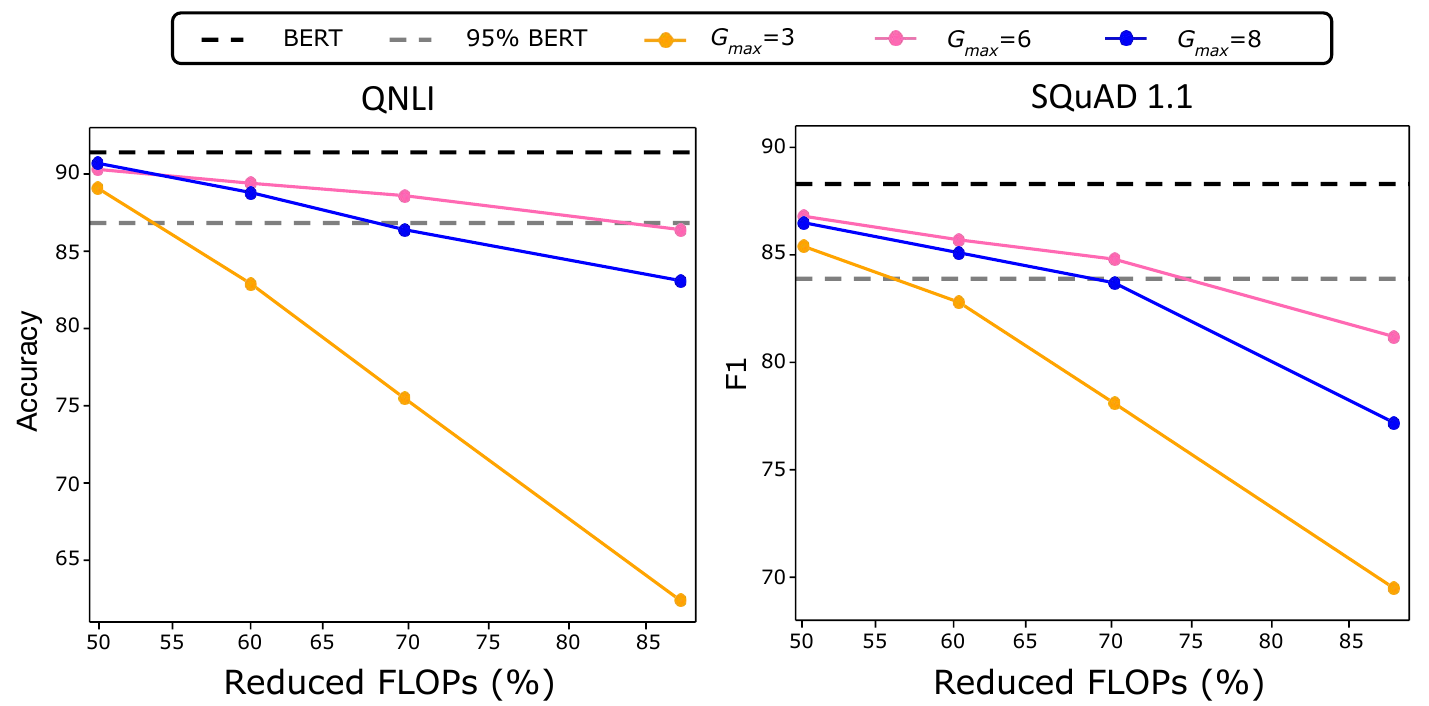}}
  
    \caption{Ablation study with respect to the number of Group ($G_{max}$) with $\tau$=1e-5}
    \label{fig:group_ablation}
\end{figure}

\begin{figure}[t!]
    \centering
    {\includegraphics[width=0.7\columnwidth]{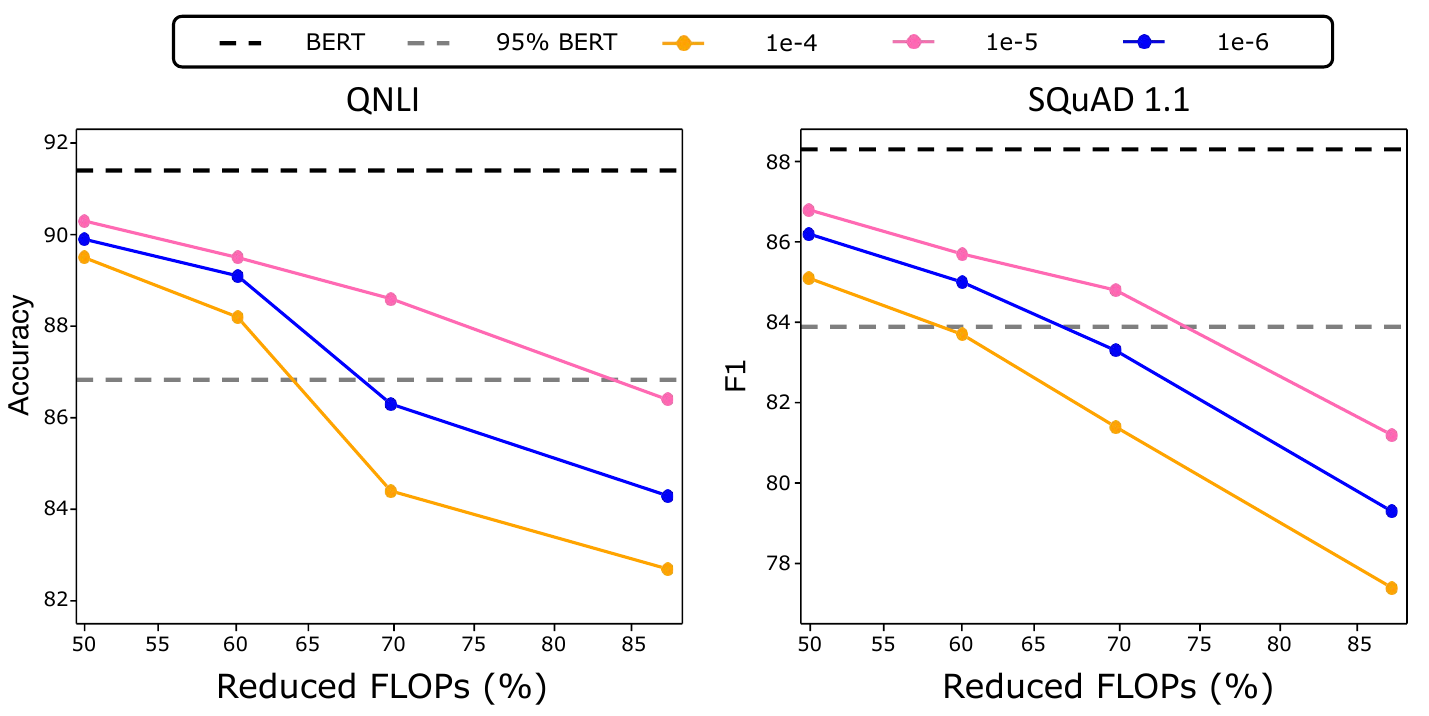}}

    \caption{Ablation study with respect to threshold ($\tau$) with $G_{max} = 6$}
    \label{fig:tau_ablation}
\end{figure}

\paragraph{Hyperparameter Sensitivity}
We conduct ablation studies to investigate the sensitivity to the hyperparameter utilized in the proposed PGB method. In Figures \ref{fig:group_ablation} and \ref{fig:tau_ablation}, we present visualization of the accuracy of the compressed model as we vary hyperparameter values, $G_{max}$ and $\tau$. 
Figure \ref{fig:group_ablation} indicates that when the number of groups is limited to a small number (i.e., ${G}_{max}=3$), there is a significant decrease in performance, which is probably because some important weights can be pruned in order to reach each target compressed size, just like in typical structured pruning. At the same time, however, it does not always mean that the larger the $G_{max}$ value, the better the final performance, as the performance also drops in the case of ${G}_{max}=8$. In terms of the threshold $\tau$, Figure \ref{fig:tau_ablation} demonstrates that there is a certain optimal point of $\tau$ to maximize the final performance. In our experiments, we found that $G_{max}=6$ and $\tau=1e-5$ produce the most promising results for compressed models.

%% file: section6.tex
\section{Conclusion}
This paper introduced PGB, one-shot semi-structured pruning with a grouping strategy, as a fast and simple compression approach for transformer-based models. PGB efficiently compresses task-specific BERT models into lightweight and accurate versions within a few hours, contrasting with other SOTA methods that take more than a day to achieve comparable results. By finding an adaptively grouped architecture, PGB combines the advantages of structured pruning and unstructured pruning, offering both computational efficiency and high accuracy. Through extensive experiments, we validated that PGB is a practical solution for quickly compressing complex transformer architectures without 
 significant performance degradation.

%% file: appendix.tex
\newpage

\setcounter{table}{0}
\setcounter{figure}{0}

\section{Experimental Details}\label{app:exdetail}
\subsection{Details for Comparison Methods}\label{app:compare}
We conduct comparative experiments on structured pruning for BERT: CoFi \cite{Xia}, BMP (based Hybrid Filled) \cite{block}, DynaBERT \cite{DynaBERT} and EBERT \cite{ebert}. All these methods belong to the iterative pruning approach, which involves performing pruning during training to improve performance. CoFi \cite{Xia} consists of 2 stages: pruning and final finetuning. Each stage involves 20 epochs of training with layer-wise distillation. BMP \cite{block} selects appropriate components to prune at the block level over 20 epochs training stages. Subsequently, prediction layer distillation is performed on the pruned model. DynaBERT \cite{DynaBERT} utilizes a 2-stage training process with knowledge distillation to dynamically prune the width and depth size, followed by a final finetuning stage. The number of training epochs varies depending on the data size. For large datasets, the width-adaptive and width- and depth-adaptive stages are performed for 1 epoch of training each, while for small datasets, each stage is performed for 3 epochs of training. Additionally, in all cases, 3 additional epochs of finetuning are conducted. EBERT \cite{ebert} involves 3 epochs of joint training during the pruning stage, followed by 3 epochs of final finetuning on the pruned model.

\subsection{Hyperparameters}\label{app:hyper}

The detailed experimental setup for PGB is provided in Table \ref{tab:hyper}. During the PGB pruning process, we utilize two hyperparameters, namely $N_{perm}$ and $G_{max}$. Also, we use the hyperparameters of the original BERT model and those of prior pruning methods for BERT.

\begin{table*} [htb]
\resizebox{\textwidth}{!}{
\begin{tabular}{c|ccccccc}
\toprule
%\noalign{\smallskip}\noalign{\smallskip}\hline
Model & \#Params & \#Encoder Layer & Hidden dim. &\#Heads & $d$ & $d_{ffn}$ \\ 
\hline

$\text{BERT}_{\text{BASE}}$ & 85M & 12& 768 &12 & 768 & 3072 \\

$\text{RoBERTa}_{\text{BASE}}$ & 100M & 12 & 768 & 12 & 768 & 3072   \\

$\text{DistilBERT}_{\text{BASE}}$ & 43M & 6 & 768 & 12 & 768 & 3072   \\

\bottomrule
\end{tabular}
}
\centering
\caption{Specifications of $\text{BERT}_{\text{BASE}}$, $\text{DistilBERT}_{\text{BASE}}$ and $\text{RoBERTa}_{\text{BASE}}$ models}
\label{tab:BERT}
\end{table*}
%-------------------------- Hyperparameters ------------------------------------
\begin{table} []
\begin{tabular}{lcc}
\noalign{\smallskip}\noalign{\smallskip}\hline
Hyperparameters &     \\
\hline
batch size  & 32 (GLUE), 16 (SQuAD) \\
pruning/finetuning epochs&   0 / 3  \\
finetuning learning rate & 2e-5, 3e-5 \\
max sequence length & 128 (GLUE), 384 (SQuAD) \\
$G_{max}$ & 6  \\
$N_{perm}$ & 6 \\
$\tau$       & 1e-5 \\    
inference batch size & 128 (GLUE) / 32 (SQuAD) \\

\hline
\end{tabular}
\centering
\caption{Hyperparameter settings of PGB in experiments}
\label{tab:hyper}
\end{table}

\begin{table}[t!]
\begin{tabular}{cccc}
\toprule
Task & \#Train &  \#Dev  & Metrics \\
\hline \hline
QNLI & 105k & 5.5k & Acc. \\
QQP  & 364k & 40k  & Acc.\\
SST-2& 67k & 872  & Acc.\\
CoLA  & 8.5k & 1k  & Matthew's corr.\\
STS-B & 7k & 1.5k  & Spearman corr.\\
MRPC  & 3.7k & 408 & Acc.\\
RTE  & 2.5k & 276 & Acc. \\
$\text{SQuAD}_{\text{v1.1}}$  & 88k & 10.5k  & EM/F1\\
$\text{SQuAD}_{\text{v2.0}}$  & 132k & 12k  & EM/F1\\
\bottomrule
\end{tabular}
\centering
\caption{Dataset summary of GLUE and $\text{SQuAD}_{\text{v1.1}}$.}
\label{GLUE&SQuAD}
\end{table}

\subsection{Details About Benchmark Datasets}\label{app:gluedetail}
 
%우리는 classification task에 해당하는 GLUE benchmark dataset에 사용하여 실험하였다. 
We perform experiments using the GLUE \cite{GLUE} and SQuAD \cite{SQuAD,SQuAD2} benchmark datasets.
%GLUE is applicable to a classification task. Among the 9 types of data in GLUE, we exclude WNLI \cite{WNLI}, which represents unstable state.  
Each task of GLUE and SQuAD datasets can fall into the following categories:
\begin{itemize}
    \item Question Answering: $\text{SQuAD}_{v1.1}$, $\text{SQuAD}_{v2.0}$
    \item Sentence Pair Similarity: QQP, MRPC \cite{MRPC}, STS-B \cite{STS-B}
    \item Natural Language Inference: RTE, QNLI
    \item Single Sentence Task: SST-2 \cite{SST-2}, CoLA \cite{CoLA}
\end{itemize}
The detailed information regarding the dataset size and metric for each task is provided in Table \ref{GLUE&SQuAD}.
% ----------------------------- GLUE --------------------------

\subsection{Model Configurations}

The experimental models used in this paper include $\text{BERT}_{\text{BASE}}$, $\text{DistilBERT}_{\text{BASE}}$ and $\text{RoBERTa}_{\text{BASE}}$ architectures, each of which has its respective parameter configuration, as described in Table \ref{tab:BERT}. The notations $d$ and $d_{ffn}$ represent the dimensionality of matrices in MHA layers and FFN layers, respectively.

\subsection{Rows and Columns for each Weight Matrix in $\text{BERT}$} 

Our PGB method performs pruning on individual parameters of $W^{Q}$, $W^{K}$, $W^{V}$, $W^{O}$ in the MHA sub-layers, as well as $W^{(1)}$ and $W^{(2)}$ in the FFN sub-layers. The number of rows and columns for each weight matrix is as follows:
\begin{itemize}
    \item ($d$, $N_{H}\times \frac{d}{N_{H}}$) : $W^{Q}$, $W^{K}$, $W^{V}$, $W^{O}$ in MHA
    \item ($d$, $d_{ffn}$), ($d_{ffn}$,\; $d$) : $W^{(1)}$, $
    W^{(2)}$ in FFN
\end{itemize}

%% file: main.bbl
\begin{thebibliography}{00}

\bibitem{Vaswani2017}
  Ashish Vaswani, Noam Shazeer, Niki Parmar, Jakob Uszkoreit, Llion Jones, Aidan N. Gomez, Lukasz Kaiser, and Illia Polosukhin,
  \textit{Attention is All you Need}, Advances in Neural Information Processing Systems, 2017, pp. 5998–6008. 

\bibitem{BERT}
  Jacob Devlin, Ming-Wei Chang, Kenton Lee, and Kristina Toutanova,
  \textit{{BERT:} Pre-training of Deep Bidirectional Transformers for Language
               Understanding},
  Proceedings of the 2019 Conference of the North American Chapter of the Association for Computational Linguistics, 2019, pp. 4171-–4186.

\bibitem{Roberta}
  Yinhan Liu, Myle Ott, Naman Goyal, Jingfei Du, Mandar Joshi, Danqi Chen, Omer Levy, Mike Lewis, Luke Zettlemoyer, and Veselin Stoyanov,
  \textit{RoBERTa: {A} Robustly Optimized {BERT} Pretraining Approach}, CoRR, abs/1907.11692, 2019.
  
\bibitem{GPT}
  Tom B. Brown, Benjamin Mann,
               Nick Ryder,
               Melanie Subbiah,
               Jared Kaplan,
               Prafulla Dhariwal,
               Arvind Neelakantan,
               Pranav Shyam,
               Girish Sastry,
               Amanda Askell,
               Sandhini Agarwal,
               Ariel Herbert{-}Voss,
               Gretchen Krueger,
               Tom Henighan,
               Rewon Child,
               Aditya Ramesh,
               Daniel M. Ziegler,
               Jeffrey Wu,
               Clemens Winter,
               Christopher Hesse,
               Mark Chen,
               Eric Sigler,
               Mateusz Litwin,
               Scott Gray,
               Benjamin Chess,
               Jack Clark,
               Christopher Berner,
               Sam McCandlish,
               Alec Radford,
               Ilya Sutskever and
               Dario Amodei,
  \textit{Language Models are Few-Shot Learners},
  Advances in Neural Information Processing Systems, 2020, pp. 1877--1901.
  
\bibitem{Han}
  Song Han, Jeff Pool, Jeff Pool, John Tran, and William J. Dally,
  \textit{Learning both Weights and Connections for Efficient Neural Network},
  Advances in Neural Information Processing Systems, 2015, pp. 1135--1143. 
  
\bibitem {KD}
Geoffrey E. Hinton, Oriol Vinyals, and Jeffrey Dean.
  \textit{Learning both Weights and Connections for Efficient Neural Network},
  Advances in Neural Information Processing Systems , CoRR, abs/1503.02531, 2015.

\bibitem {SNIP}
  Lee, Namhoon, Thalaiyasingam Ajanthan, and Philip HS Torr,
  \textit{Snip: Single-shot network pruning based on connection sensitivity},
  arXiv preprint arXiv:1810.02340, 2018. 

\bibitem {Thinet}
  Jian-Hao Luo, Jianxin Wu, and Weiyao Lin,
  \textit{Thinet: A filter level pruning method for deep neural network compression},
  In Proceedings of the IEEE international conference on computer vision, 2017, pp. 5058--5066. 

\bibitem {LoB}
  Tianlong Chen, Jonathan Frankle, Shiyu Chang, Sijia Liu, Yang Zhang, Zhangyang Wang, and Michael Carbin,
  \textit{The lottery ticket hypothesis for pretrained bert networks},
  Advances in neural information processing systems, 2020, 33:pp. 15834-15846.

\bibitem {DynaBERT}
  Lu Hou, Zhiqi Huang, Lifeng Shang, Xin Jiang, Xiao Chen, and Qun Li,
  \textit{Dynabert: Dynamic BERT with adaptive width and depth},
  In Advances in neural information processing systems, 2020, 33.

\bibitem {block}
  Fran{\c{c}}ois Lagunas, Ella Charlaix, Victor Sanh, and Alexander M. Rush,
  \textit{Block Pruning For Faster Transformers},
  In Empirical Methods in Natural Language Processing (EMNLP), 2021, pp. 10619--10629.

\bibitem {Xia}
  Mengzhou Xia, Zexuan Zhong, and Danqi Chen,
  \textit{Structured Pruning Learns Compact and Accurate Models},
  In Proceedings of the 60th Annual Meeting of the Association for Computational Linguistics (Volume 1: Long Papers), 2022, pp. 1513--1528.

\bibitem {DistilB}
  Victor Sanh, Lysandre Debut, Julien Chaumond, and Thomas Wolf,
  \textit{Structured Pruning Learns Compact and Accurate Models}, CoRR, abs/1910.01108, 2019.

\bibitem {TinyBERT}
  Xiaoqi Jiao , Yichun Yin, Lifeng Shang, Xin Jiang, Xiao Chen, Linlin Li, Fang Wang and Qun Liu,
  \textit{Structured Pruning Learns Compact and Accurate Models}, CoRR, abs/1910.01108, 2019.

\bibitem {DGC}
  Zhuo Su, Linpu Fang, Wenxiong Kang, Dewen Hu Matti Pietikäinen, and Li Liu,
  \textit{Dynamic group convolution for accelerating convolutional neural networks}, In Computer Vision–ECCV 2020: 16th European Conference, 2020, pp. 138--155.

\bibitem {Zhao}
  Ruizhe Zhao and Wayne Luk,
  \textit{Efficient structured pruning and architecture searching for group convolution}, In Proceedings of the IEEE/CVF International Conference on Computer Vision Workshops, 2019, pp. 1961--1970.

\bibitem {GroupFormer}
Sungrae Park, Geewook Kim, Junyeop Lee, Junbum Cha, Ji-Hoon Kim, and Hwalsuk Lee,
\textit{Scale down Transformer by Grouping Features for a Lightweight Character-level Language Model}, Proceedings of the 28th International Conference on Computational Linguistics, 2020, pp. 6883–-6893.

\bibitem {groupbert}
Ivan Chelombiev, Daniel Justus, Douglas Orr, Anastasia Dietrich, Frithjof Gressmann, Alexandros Koliousis, and Carlo Luschi,
\textit{Groupbert: Enhanced transformer architecture with efficient grouped structured}, arXiv preprint arXiv:2106.05822, 2021.

\bibitem {GLUE}
Alex Wang, Amanpreet Singh, Julian Michael, Felix Hill, Omer Levy, and Samuel R. Bowman,
\textit{{GLUE:} {A} Multi-Task Benchmark and Analysis Platform for Natural
               Language Understanding}, In International Conference on Learning Representations (ICLR), 2019.
               
\bibitem{SQuAD}
Pranav Rajpurkar, Jian Zhang, Konstantin Lopyrev, and Percy Liang,
\textit{SQuAD: 100, 000+ Questions for Machine Comprehension of Text}, In Empirical Methods in Natural Language Processing (EMNLP), 2016.


\bibitem {MiniLM}
Wenhui Wang, Furu Wei, Li Dong, Hangbo Bao, Nan Yang, and Ming Zhou,
\textit{MiniLM: Deep Self-Attention Distillation for Task-Agnostic Compression
               of Pre-Trained Transformers}, In Advances in Neural Information Processing Systems, 2020, pp. 5776--5788.

\bibitem {MoB}
Zhiqing Sun, Hongkun Yu, Xiaodan Song, Renjie Liu, Yiming Yang, and Denny Zhou,
\textit{MobileBERT: a Compact Task-Agnostic {BERT} for Resource-Limited Devices}, In Proceedings of the 58th Annual Meeting of the Association for Computational Linguistics (ACL), 2020, pp. 2158--2170.

\bibitem {Mov}
Victor Sanh, Thomas Wolf, and Alexander Rush,
\textit{Movement pruning: Adaptive sparsity by fine-tuning}, Advances in Neural Information Processing Systems, 2020, 33: pp.20378--20389.

\bibitem {sixteen}
Paul Michel, Omer Levy, and Graham Neubig,
\textit{Are sixteen heads really better than one?}, Advances in Neural Information Processing Systems (Volume 32), 2019.

\bibitem {voita}
Elena Voita, David Talbot, Fedor Moiseev, Rico Sennrich, and Ivan Titov,
\textit{Analyzing multi-head self-attention: Specialized heads do the heavy lifting, the rest can be pruned}, arXiv preprint arXiv:1905.09418, 2019.

\bibitem {SMP}
Zhengyan Zhang, Fanchao Qi, Zhiyuan Liu, Qun Liu, and Maosong Sun,
\textit{Know what you don't need: Single-Shot Meta-Pruning for attention heads}, {AI} Open, 2021, 2: pp. 36--42.

\bibitem {layer1}
Angela Fan, Edouard Grave, and Armand Joulin,
\textit{Reducing Transformer Depth on Demand with Structured Dropout}, CoRR, abs/1909.11556, 2019.

\bibitem {layer2}
Hassan Sajjad, Fahim Dalvi, Nadir Durrani, and Preslav Nakov,
\textit{Poor Man's {BERT:} Smaller and Faster Transformer Models}, CoRR, abs/2004.03844, 2020.

\bibitem {earlybert}
Xiaohan Chen, Yu Cheng, Shuohang Wang, Zhe Gan, Zhangyang Wang, and Jingjing Liu,
\textit{Earlybert: Efficient bert training via early-bird lottery tickets}, arXiv preprint arXiv:2101.00063, 2020.

\bibitem {deeproot}
Yani Ioannou, Duncan Robertson, Roberto Cipolla, and Antonio Criminisi,
\textit{Deep roots: Improving cnn efficiency with hierarchical filter groups}, Proceedings of the IEEE conference on computer vision and pattern recognition, 2017, pp. 1231--1240.

\bibitem {xie}
Guotian Xie, Jingdong Wang, Ting Zhang, Jianhuang, Lai, Richang Hong, and Guo-Jun Qi,
\textit{Interleaved structured sparse convolutional neural networks}, In Proceedings of the IEEE Conference on Computer Vision and Pattern Recognition, 2018, pp. 8847--8856.

\bibitem {Brain}
Babak Hassibi, David G. Stork, and Gregory J. Wolff,
\textit{Interleaved structured sparse convolutional neural networks}, In Proceedings of International Conference on Neural Networks (ICNN'88), San Francisco, CA, USA, March 28 - April 1, 1993, pp. 293--299.

\bibitem {second-order}
Pavlo Molchanov, Arun Mallya, Stephen Tyree, Iuri Frosio, and Jan Kautz ,
\textit{Importance Estimation for Neural Network Pruning}, In {IEEE} Conference on Computer Vision and Pattern Recognition, {CVPR}
                  2019, Long Beach, CA, USA, June 16-20,2019, pp. 11264--11272.

\bibitem {WoodFisher}
Sidak Pal Singh and Dan Alistarh,
\textit{WoodFisher: Efficient Second-Order Approximation for Neural Network
                  Compression}, In Advances in Neural Information Processing Systems 33: Annual Conference
                  on Neural Information Processing Systems 2020, NeurIPS 2020, December
                  6-12, 2020.

\bibitem {SQuAD2}
Rajpurkar, Pranav and Jia, Robin and Liang, Percy,
\textit{Know what you don't know: Unanswerable questions for SQuAD}, arXiv preprint arXiv:1806.03822, 2018.

\bibitem {Pytorch}
Adam Paszke, Sam Gross, Francisco Massa, Adam Lerer, James Bradbury, Gregory Chanan, Trevor Killeen, Zeming Lin, Natalia Gimelshein, Luca Antiga, Alban Desmaison, Andreas K{\"{o}}pf, Edward Z. Yang, Zach DeVito, Martin Raison, Alykhan Tejani, Sasank Chilamkurthy,  Benoit Steiner, Lu Fang, Junjie Bai, and
                  Soumith Chintala,
\textit{PyTorch: An Imperative Style, High-Performance Deep Learning Library}, CoRR, abs/1912.01703, 2019.

\bibitem {Wolf}
Thomas Wolf, Lysandre Debut, Victor Sanh, Julien, Chaumond, Clement Delangue, Anthony Moi, Pierric  Cistac, Tim Rault, R{\'e}mi Louf, Morgan Funtowicz,
\textit{Transformers: State-of-the-art natural language processing}, In Proceedings of the 2020 conference on empirical methods in natural language processing: system demonstrations, 2020, pp. 38--45.

\bibitem {ebert}
Liu, Zejian and Li, Fanrong and Li, Gang and Cheng, Jian,
\textit{EBERT: Efficient BERT inference with dynamic structured pruning}, Findings of the Association for Computational Linguistics: ACL-IJCNLP, 2021, pp. 4814--4823.

\bibitem {TFsnip}
Zi Lin, Jeremiah Z. Liu, Zi Yang, Nan Hua, and Dan Roth,
\textit{Pruning Redundant Mappings in Transformer Models via Spectral-Normalized
                  Identity Prior}, Findings of the Association for Computational Linguistics: {EMNLP}
                  2020, Online Event, 16-20 November, 2020, pp. 719--730.

\bibitem {shufflenet}
Zhang, Xiangyu and Zhou, Xinyu and Lin, Mengxiao and Sun, Jian,
\textit{Shufflenet: An extremely efficient convolutional neural network for mobile devices}, In Proceedings of the IEEE conference on computer vision and pattern recognition, 2018, pp. 6848--6856.

\bibitem {platon}
Qingru Zhang, Simiao Zuo, Chen Liang, Alexander Bukharin, Pengcheng He, Weizhu Chen, and Tuo Zhao,
\textit{Platon: Pruning large transformer models with upper confidence bound of weight importance}, In Proceedings of the IEEE conference International Conference on Machine Learning, 2022, pp. 26809--26823.

\bibitem {SST-2}
Richard Socher, Alex Perelygin, Jean Wu, Jason Chuang, Christopher D. Manning, Andrew Y. Ng, and Christopher Potts,
\textit{Recursive Deep Models for Semantic Compositionality Over a Sentiment Treebank}, In Proceedings of the 2013 Conference on Empirical Methods in Natural Language Processing (EMNLP), 2013, pp. 1631--1642.

\bibitem {CoLA}
Alex Warstadt, Amanpreet Singh, and Samuel R. Bowman,
\textit{Neural Network Acceptability Judgments}, Transactions of the Association of Computational
Linguistics (TACL), 2019, 7: pp. 625--641.

\bibitem {STS-B}
Daniel M. Cer, Mona T. Diab, Eneko Agirre, I{\~{n}}igo Lopez{-}Gazpio, and Lucia Specia,
\textit{SemEval-2017 Task 1: Semantic Textual Similarity Multilingual and
               Crosslingual Focused Evaluation}, In Proceedings of the 11th International Workshop on Semantic Evaluation (SemEval-2017), 2017, 7: pp.1--14.

\bibitem {MRPC}
William B. Dolan and Chris Brockett, I{\~{n}}igo Lopez{-}Gazpio, and Lucia Specia,
\textit{Automatically Constructing a Corpus of Sentential Paraphrases}, In Proceedings of the Third International Workshop on Paraphrasing, IWP@IJCNLP 2005, 2005.            
             

\end{thebibliography}
